# Phase Transition of Tractability in Constraint Satisfaction and Bayesian Network Inference

Yong Gao
Department of Computing Science
University of Alberta
Edmonton, Canada T6G 2E8
ygao@cs.ualberta.ca

## Abstract

Identifying tractable subclasses and designing efficient algorithms for these tractable classes are important topics in the study of constraint satisfaction and Bayesian network inference problems. In this paper we investigate the asymptotic average behavior of a typical tractable subclass characterized by the treewidth of the problems. We show that the property of having a bounded treewidth in the constraint satisfaction problem and Bayesian network inference problem has a phase transition that occurs while the underlying structures of problems are still sparse. This implies that algorithms making use of treewidth based structural knowledge only work efficiently in a limited range of random instances.

## 1 INTRODUCTION

It is well known that many NP complete problems have tractable subclasses characterized by certain structural parameters. The treewidth is one of such parameters and has drawn much attention in algorithmic graph theory [1, 2] and artificial intelligence [3].

In the study of constrain satisfaction problems (CSPs) and Bayesian network inference, there has been much effort in designing efficient algorithms that make best use of the property of bounded treewidth of the problems. The notion of tractable classes of CSPs parameterized by treewidth can be traced back to the work of [4] and since then, has remained an interesting topic [3, 5, 6, 7]. CSPs with a bounded treewidth can be solved polynomially using dynamic programming techniques. For Bayesian networks with a tree structure, the famous message-passing algorithm solves the inference problem in linear time [8]. For Bayesian networks with arbitrary structures, the most widely used method is that of join-tree which transforms the original inference problem into one on a tree of subsets of variables. The transformation is based on triangulation and tree decomposition of the given networks. As the size of subsets in the tree decomposition is directly related to the time and space complexity of the join-tree algorithm, there has been much work on finding the optimal decomposition, which itself is also an NP complete problem. See, for example, the work of [9, 10, 11] for more details. Another recently proposed approach is to make sure that the Bayesain networks have a controlled treewidth when constructing them [12, 13].

Phase transitions and threshold phenomena in random graphs and combinatorial search problems have been extensively studied during the past decade [14, 15]. A phase transition in combinatorial search refers to the phenomenon that the probability that a random instance of the problem has a solution drops abruptly from one to zero as some order parameters of the random model crosses a critical value called the *threshold*. Associated with this phase transition in solubility is the dramatic change of the hardness of a problem, and the hardest instances usually occur at the phase transition.

It is now a common practice to generate random instances of combinatorial search problems at the phase transition as benchmarks [16]. Randomly generated Bayesian networks have also been widely used in the evaluation and comparison of different inference algorithms [17, 18].

In this paper, we study the treewidth-based tractability of CSPs and Bayesian network inference by investigating the asymptotic probabilistic behavior of the property of having a bounded treewidth. Working on models of random constraint satisfaction problems and random Bayesian networks, we show that the bounded treewidth-based tractability has a phase transition which occurs when the underlying structures of the problems are still quite sparse.

In section 2, we briefly review CSPs, Bayesian networks, and their random models. In section 3, we introduce two random graph models. One is the classical model of random graphs and another is termed as the graph of random cliques and is specially designed for modelling CSPs and Bayesian networks. In sections 4 and 5, we study the phase transition of the bounded treewidth property for graphs of random cliques, and apply the results to random models of CSPs and Bayesian networks. We conclude in section 6 with some discussions about the implication of our results and future work.

## 2 CONSTRAINT SATISFACTION PROBLEMS AND BAYESIAN NETWORKS

In this section, we briefly review the terminology in constraint satisfaction problems and Bayesian networks, introduce their random models, and discuss the treewidth-based tractability and algorithms.

### 2.1 CONSTRAINT SATISFACTION PROBLEMS AND THEIR RANDOM MODELS

A constraint satisfaction problem $\mathcal{C} = (X, D, C)$ consists of a set of variables $X = (x_1, \cdots, x_n)$, a domain $D$ for the variables, and a set of constraints $C = (C_1, \cdots, C_m)$. Each constraint $C_i$ is specified by its scope, a subset of $X$, and a set of restrictions on the scope variables. We will misuse the notation to denote by $|C_i|$ the size of the scope of the constraint and assume throughout this paper that $|C_i| = d$ is fixed for each $i$ and call $d$ the order of the CSP.

The *primal graph* of a CSP $\mathcal{C} = (X, D, C)$ is a graph $G = G(V, E)$ where $V$ corresponds to the set of variables $X$ and $(v_i, v_j) \in E$ if and only if the corresponding variables $x_i$ and $x_j$ appear in some constraints at the same time. A CSP also has a natural hypergraph representation where the subsets of variables of the constraints are treated as hyperedges.

A random constraint problem, denoted by $\mathcal{C}(n, m, \mathcal{P})$, is obtained by first randomly selecting $m$ size-$d$ subsets of variables and then choosing a constraint for each subset of variables independently according to the probability distribution $\mathcal{P}$. For the purpose of the current paper, we do not need to consider the specific properties of the probability distribution $\mathcal{P}$, and therefore, will simply write the random CSP as $\mathcal{C}(n, m)$.

The constraint satisfaction problem is among the search problems in which the phase transition in solubility has been identified and extensively investigated. We refer interested readers to the work of [19, 20] and the references therein.

### 2.2 BAYESIAN NETWORKS AND THEIR RANDOM MODELS

Given a set of random variables $X = (X_1, \cdots, X_n)$, a Bayesian network is a pair $\mathcal{B}(G, P)$ where $G$ is a directed acyclic graph over the set of nodes $X$ and $P$ defines a set of conditional probabilities $P_i = Pr\{X_i|pa(X_i)\}$ with $pa(X_i)$ being the parent of the node $X_i$. A Bayesian network provides a concise representation of the probability distribution of the random vector $X$. The *moral graph* of a Bayesian network is an undirected graph obtained by first connecting the parents of each nodes, and then changing the directed edges into undirected ones.

To study the average behavior of the structural properties of a Bayesian networks, we introduce a random Bayesain network model as follows.

**Definition 2.1** *Given a set of random variables $X = (X_1, \cdots, X_n)$, a random Bayesian network $\mathcal{B}(n)$ is specified by selecting the parents of each node randomly and independently. If we assume that the node $X_i$ chooses as its parent each of the rest of the nodes randomly and independently with the probability $p_i$, we use $\mathcal{B}(n, (p_i, 1 \leq i \leq n))$ to denote the corresponding random model.*

Of course, the above random model is not guaranteed to generate directed acyclic graphs. To generate directed acyclic networks, we may consider a modified version of the model that first chooses a random order of the variables, and then lets each variable select their parents from the precedent variables according the order. The idea of our analysis can be extended to this restricted model with some complication.

### 2.3 TREEWIDTH AND TREEWIDTH BASED TRACTABILITY

The concepts of treewidth and tree-decomposition on graphs generalize those of trees. The treewidth of a graph is usually defined in two equivalent ways.

**Definition 2.2** *[2] k-Trees are defined recursively as follows:*

1. *A clique with $k+1$ vertices is a k-tree;*

2. *Given a k-tree $T_n$ with $n$ vertices, a k-tree with $n + 1$ vertices is constructed by adding to $T_n$ a new vertex which is made adjacent to a k-clique of $T_n$ and non-adjacent to rest of the vertices.*

**Definition 2.3** *[2] A graph is called a partial k-tree if it is a subgraph of a k-tree. The treewidth of a graph G is the minimum value k for which G is a partial k-tree.*

Equivalently, the treewidth of a graph can be defined using the concept of tree decomposition of a graph.

**Definition 2.4** *[2] A tree decomposition of a graph $G = (V, E)$ is a pair $D = (S, T)$ where $S = \{X_i, i \in I\}$ is a collection of subsets of vertices of $G$ and $T = (I, F)$ is a tree with one node for each subset of $S$, such that*

1. $\bigcup_{i \in I} X_i = V$,

2. *for all the edges $(v, w) \in E$ there exists a subset $X_i \in S$ such that both $v$ and $w$ are in $X_i$, and*

3. *for each vertex the set of nodes $\{i, x \in X_i\}$ forms a subtree of $T$.*

*The width of the tree decomposition $D = (S, T)$ is $max_{i \in I}(|X_i| - 1)$. And the treewidth of a graph is the minimum width over all tree decompositions of the graph.*

A detailed discussion on treewidth, related algorithms, and applications can be found in [2]. Of particular interest is the following result on the treewidth of a classical random graph, which serves as a starting point of the current paper.

**Lemma 2.1** *[2] Let $\delta \geq 1.18$. Then a random graph $G(n, m)$ (See next section for the definition) with $m \geq \delta n$ almost surely has treewidth $\Theta(n)$.*

Many NP hard problems can be solved polynomially when restricted to instances with a bounded treewidth [1]. When restricted to instances whose underlying graphs have a bounded treewidth, both the CSP and Bayesian network inference problem are tractable [5, 3]. The basic idea is to decompose the original problems into a set of subproblems so that the subproblems can be represented as a tree. For CSPs, solutions to the original problems can be found by solving this tree of subproblems using the idea of dynamic programming. For the Bayesian networks, the original inference problem can be solved by carrying out the probabilistic calculation on the clique tree. In both CSP and Bayesian network inference, the subproblem decomposition is based on finding the tree decomposition of the underlying graphs of the problems, and the algorithm is time and space exponential in the maximal size of the subproblems.

## 3 RANDOM GRAPHS AND GRAPHS OF RANDOM CLIQUES

The theory of random graphs was founded by Erdös and Rényi in 1960s and since then, has been an intensively studied topic. There are two basic models of random graphs $G(n, m)$ and $G(n, p)$. In the $G(n, m)$ model, the $m$ edges are chosen randomly without replacement from the set of all the pairs of the vertices, while in the $G(n, p)$ model, each pair of vertices is selected as an edge independently and randomly with the probability $p$. Over the years, it has been shown that many interesting graph properties have a threshold phenomenon, i.e., there exists a threshold function such that when the number of the edges or the edge probability as a function of $n$ increase faster than the threshold function, asymptotically almost surely the random graph has the property [14]. An interesting connection between the threshold phenomenon of random graphs and the hardness of many NP complete search problems such as SAT, CSPs, and graph coloring has been established over the past ten years [15, 21, 19].

As a generalization to random graphs, we can also consider random hypergraphs [22]. The most commonly used model of random hypergraph is $G^d(n, m)$ in which the $m$ hyper-edges are chosen randomly without replacement from the set of all the hyper-edges of size $d$.

In the following, we introduce a new random model of graphs which can be viewed as an abstraction of the "random" graphs generated from the random model of CSPs and Bayesian networks discussed in the previous section. We call the model graph of random cliques.

**Definition 3.1** *Given a vertex set $V$, a graph of random cliques $G_C(n, m)$ is a graph obtained from a random hypergraph by making adjacent all the pairs of vertices that belongs to a hyperedge. A graph of random cliques is called d-uniform if it is obtained from a d-uniform random hypergraph $G^d(n, m)$, and is denoted by $G_C^d(n, m)$.*

For example, the primal graph of a random CSP is a graph of random cliques and the classical random graph $G(n, m)$ is a 2-uniform graph of random cliques.

## 4 TREEWIDTH OF RANDOM CSPS

In this section, we study the treewidth of random CSPs. We first prove a result on the treewidth of graphs of random cliques. Let $G_C^d(n, m)$ be a $d$-uniform graph of random cliques. Denote by $\mathcal{W}$ the graph prop-erty of having an $o(n)$ treewidth.

**Definition 4.1** [2] *Let $G(V,E)$ be a graph with $|V| = n$. A partition $(S,A,B)$ of $V$ is a balanced $k$-partition if the following conditions are satisfied:*

1. $|S| = k+1$;

2. $\frac{1}{3}(n-k-1) \leq |A|, |B| \leq \frac{2}{3}(n-k-1)$; and

3. $S$ separates $A$ and $B$, i.e., there are no edges between vertices of $A$ and vertices of $B$.

**Theorem 4.1** *Let $c = \frac{\ln 2}{d \ln 3 - \ln(1+2^d)}$.*

$$\lim_n Pr\{G_{\mathcal{C}}^d(n,m) \in \mathcal{W}\} = \begin{cases} 1, & \text{if } \frac{m}{n} < \frac{1}{d(d-1)}; \\ 0, & \text{if } \frac{m}{n} > c. \end{cases} \quad (1)$$

Proof: If $\frac{m}{n} < \frac{1}{d(d-1)}$, then the random hypergraph $G^d(n,m)$ almost surely contains hypertrees and unicycles [22]. It can be shown that the graph of cliques obtained from a hypergraph with only hypertrees and unicycles has a treewidth of at most $d+1$.

Now we consider the case of $\frac{m}{n} > c$. Let $w(n,m,d)$ be the treewidth of $G_{\mathcal{C}}^d(n,m)$. We claim that there exists a $0 < \delta < 1$ such that

$$\lim_n Pr\{w(n,m,d) \leq \delta n\} = 0. \quad (2)$$

The proof is a generalization to that of [2]. It is well-known that if a graph has a treewidth less than or equal to $k$, then the graph must have a balanced $k$-partition [2]. Let $\mathcal{P}$ be the set of all the $k$-partitions of the vertex set $V$ that satisfies the first two conditions of the definition of balanced partition. For a given $P = (S,A,B) \in \mathcal{P}$, define a random variable $I_P$ as follows:

$$I_P = \begin{cases} 1, & \text{if } P \text{ is a balanced partition}; \\ 0, & \text{otherwise}. \end{cases}$$

It is easy to see that $I_P = 1$ if and only if there are no edges between vertices in $A$ and vertices in $B$ in the graph of random cliques $G_{\mathcal{C}}^d(n,m)$.

Let $N = \binom{n}{d}$ be the number of possible hyperedges of size $d$ and $N_P$ the number of possible hyperedges that are subsets of either $A \bigcup S$ or $B \bigcup S$. Let $a = |A|$, we have

$$N_P = \binom{a+k+1}{d} + \binom{n-a}{d} - \binom{k+1}{d}$$

$$\leq \binom{a+k+1}{d} + \binom{n-a}{d}, \quad (3)$$

and thus,

$$\frac{N_p}{N} \leq \frac{(a+k+1)\cdots(a+k+1-d+1)}{n(n-1)\cdots(n-d+1)}$$
$$+ \frac{(n-a)\cdots(n-a-d+1)}{n(n-1)\cdots(n-d+1)}$$
$$\leq \left(\frac{a+k+1}{n}\right)^d + \left(\frac{n-a}{n}\right)^d$$
$$= \frac{(a+k+1)^d + (n-a)^d}{n^d}. \quad (4)$$

Write $y = \frac{k+1}{n}$ and consider the function $f(a) = (a+k+1)^d + (n-a)^d$ defined on the interval

$$[\frac{1}{3}(n-k-1), \frac{2}{3}(n-k-1)].$$

It is easy to see that $f'(a) = 0$ at $a = \frac{1}{2}n(1-y)$. If $y = \frac{k+1}{n}$ is sufficient small, then, $f(a)$ is maximized at $a = \frac{1}{3}(n-k-1)$. Therefore, we have

$$\frac{N_p}{N} \leq \frac{1}{n^d}\left((\frac{1}{3}n+k+1)^d + (\frac{2}{3}n+k+1)^d\right)$$
$$= (\frac{1}{3})^d(1+3y)^d + (\frac{2}{3})^d(1+\frac{3}{2}y)^d$$
$$\leq \left((\frac{1}{3})^d + (\frac{2}{3})^d\right)(1+3y)^d \quad (5)$$

It follows that

$$E\{I_P\} \leq \frac{\binom{N_P}{m}}{\binom{N}{m}} \leq \left(\left(\frac{1}{3}\right)^d + \left(\frac{2}{3}\right)^d\right)^m (1+3y)^{dm}.$$

Let $I = \sum_{P \in \mathcal{P}} I_P$. By its definition, we have

$$|\mathcal{P}| = \binom{n}{k+1} \sum_{\frac{1}{3}(n-k-1) \leq a \leq \frac{2}{3}(n-k-1)} \binom{n-k-1}{a}$$
$$\leq \binom{n}{k+1} 2^n.$$

It follows that the expectation of $I$ satisfies

$$E\{I\} = \sum_{P \in \mathcal{P}} E\{I_P\}$$
$$\leq \binom{n}{k+1} 2^n \left(\left(\frac{1}{3}\right)^d + \left(\frac{2}{3}\right)^d\right)^m (1+3y)^{dm}.$$

Recall that $0 < y = \frac{k+1}{n} < 1$. We obtain from Stirling's formula that

$$\binom{n}{k+1} \sim \frac{1}{\sqrt{2\pi y(1-y)n}} \left(\frac{1}{y^y(1-y)^{1-y}}\right)^n.$$

And hence,

$$E\{I\} \leq \frac{1}{\sqrt{2\pi y(1-y)n}} \left(\frac{2}{y^y(1-y)^{1-y}}\right)^n$$
$$\cdot \left(\left(\frac{1}{3}\right)^d + \left(\frac{2}{3}\right)^d\right)^m (1+3y)^{dm}. \quad (6)$$

Notice that

$$\lim_{y \to 0} \frac{2}{y^y(1-y)^{1-y}} = 2.$$

For any $\frac{m}{n} > c$ with $c$ satisfying

$$\left(\left(\frac{1}{3}\right)^d + \left(\frac{2}{3}\right)^d\right)^c < \frac{1}{2},$$

let $y = \frac{k+1}{n}$ be small enough so that

$$\frac{2}{y^y(1-y)^{1-y}} \left(\left(\frac{1}{3}\right)^d + \left(\frac{2}{3}\right)^d\right)^c (1+3y)^{dc} < 1$$

and let $\delta = y$, we have

$$\lim_n Pr\{w(n,m,d) \leq \delta n\} \leq \lim_n Pr\{I > 0\}$$
$$\leq \lim_n E[I] = 0,$$

that is, (2) is true. The theorem is proved.

By applying the above theorem to the primal graph of the random CSP, we get

**Corollary 4.1** *Let $C(n,m)$ be a random CSP. Then with probability asymptotic to one, the running time of any tree-decomposition based algorithms is exponential in $n$ if $\frac{m}{n} > \frac{\ln 2}{d \ln 3 - \ln(1+2^d)}$.*

## 5 TREEWIDTH OF BAYESIAN NETWORKS

We start our discussion by first considering the simpler case of two-layer Bayesian networks which can be represented as a directed bipartite graph. A typical example is the QMR-DT database where the upper layer has about 600 nodes representing diseases and the lower layer has about 4000 nodes representing the symptoms [23]. Even with such a simple structure, the exact inference generally remains intractable. See [23] for empirical evidence and [24] for an idea of an NP-complete proof.

For two layer Bayesian networks, we may consider the following straightforward random model $\mathcal{B}(V_1, V_2, d)$ where $V_1$ and $V_2$ are respectively the sets of nodes of upper and lower layers, and each node $x \in V_2$ randomly chooses a set of $d$ nodes in $V_1$ as its parents.

**Theorem 5.1** *Let $\mathcal{B}(V_1, V_2, d)$ be a random two-layer Bayesian network with $|V_1| = n$, $|V_2| = m$ and treewidth $w(m,n)$. Let $\mathcal{W}$ be the property that $w(m,n) \in o(n)$ and let $c = \frac{\ln 2}{d \ln 3 - \ln(1+2^d)}$. Then, we have*

$$\lim_n Pr\{\mathcal{B}(V_1, V_2, d) \in \mathcal{W}\}$$
$$= \begin{cases} 1, & \text{if } \frac{m}{n} < \frac{1}{d(d-1)}; \\ 0, & \text{if } \frac{m}{n} > c. \end{cases} \quad (7)$$

*Proof:* Let $G(V_1, V_2)$ be the moral graph of the Bayesian network and $G_1(V_1)$ be the induced graph of $G(V_1, V_2)$. By the definition of the treewidth and the fact that $G(V_1, V_2)$ is bipartite, it can be shown that the treewidth of $G(V_1, V_2)$ is the maximum of $d+1$ and the treewidth of $G_1(V_1)$. The theorem is proved by applying Theorem 4.1 to the graph of random cliques $G_1(V_1)$.

For general Bayesian networks, if we want to use theorem 4.1, then the cliques in their *moral graph* have to be added randomly and independently. This is however not an appropriate assumption in the context of Bayesian networks because (1) the generated networks are not guaranteed to be acyclic and (2) there is no reason to assume that each variable has the same constant number of parents. The random model introduced in Definition 2.1 is a first step toward a more realistic random model for Bayesian networks, where we assume that each node selects its parents randomly and independently. It should be noted that this model can still generate cyclic networks. However, the idea of the analysis on this model can be extended to more elaborated models with some complication.

**Theorem 5.2** *Let $\mathcal{B}(n, (p_i, 1 \leq i \leq n))$ be a random Bayesian networks on $n$ variables and $w(n)$ the treewidth of its moral graph. Then, there exists a $0 < \delta < 1$ such that $\lim_n Pr\{w(n) \leq \delta n\} = 0$ if*

$$(\prod_{i=1}^n (1-p_i))^{\frac{1}{3}} < \frac{1}{2}.$$

*Proof:* Similar to the proof of theorem 4.1, let $\mathcal{P}$ be the set of all the $k-$partitions of the vertex set of the moral graph of the Bayesian network that satisfies the first two conditions of the definition of balanced partition. For a given $P = (S, A, B) \in \mathcal{P}$, let $E$ be the event that $P$ is a balanced partition, i.e., the event that there are no edges between vertices of $A$ and vertices of $B$.

For each $1 \leq i \leq n$ with $X_i \in A$ (or $X_i \in B$), let $E_i$ be the event that all of its parents are in $A \bigcup S$ (or in $B \bigcup S$ respectively). For $X_i \in S$, let $E_i$ be the event that all of its parents are in $A \bigcup S$ or in $B \bigcup S$. We

have

$$E = \bigcap_{1 \leq i \leq n} E_i.$$

Since by assumption, each node selects its parents independently from the others, we have

$$Pr\{E\} = \prod_{i=1}^{n} Pr\{E_i\}. \tag{8}$$

For $X_i \in A$ (or $X_i \in B$), we have

$$Pr\{E_i\} \leq (1-p_i)^{\frac{1}{3}(n-k-1)}$$

and for $X_i \in S$, we have

$$Pr\{E_i\} \leq 2(1-p_i)^{\frac{1}{3}(n-k-1)} - (1-p_i)^k.$$

The rest of proof is similar to that of theorem 4.1.

## 6  CONCLUSIONS AND FUTURE WORK

CSPs and Bayesian networks with a bounded treewidth are the most studied tractable subclasses in constraint programming and Bayesian network inference. In this paper, we have shown that in both of the problems, the property of having a bounded treewidth has a phase transition that occurs while the underlying structures of the problems are still quite sparse. It would be interesting to know the behavior of the treewidth inside the interval of the phase transition.

From the random graph theory, the edge-to-vertex ratio $\frac{m}{n} = \frac{1}{2}$ is the threshold for the appearance of the "giant component" of order $n$ in the random graph $G(m,n)$. Below this threshold, a random graph almost always contains only tree and unicycle components, and hence, has a treewidth of at most 3. Above this threshold, a random graph almost always contains a "giant" connected component. $\frac{m}{n} = \frac{1}{2}$ is also known to be the threshold for the property of being planar. We believe and are working to show that with probability asymptotic to 1, a random graph with an edge-to-vertex ratio above the threshold $\frac{1}{2}$ cannot have a fixed constant treewidth. If this is the case, then the classical random graph model is surely not an interesting one for the modelling purpose of exact algorithms that depend on the property of bounded-treewidth.

In the study of Bayesian network inference, randomly generated networks have been widely used to evaluate and compare various inference algorithms [17, 18]. Our results show that the treewidth of the random instances is asymptotically in the order of the size of the networks even if the random model is quite sparse. This implies that purely random Bayesian networks are not adequate at least for the evaluation of tree-decomposition based inference algorithms. A natural question then is how to devise a random model that has a controlled treewidth. Motivated by the k-tree based definition of treewidth, we propose the following random model. Starting from a clique of k nodes, we add new nodes one at a time. The new node is then connected to the nodes of a randomly selected k-clique in the old graph. We illustrate the idea by giving the following random Bayesian network model.

**Definition 6.1** Let $X = (X_1, X_2, \cdots, X_n)$ be a random vector. A random Bayesian network with bounded treewidth(RBNBT) is a Bayesian network constructed using the following procedure

1. Randomly select k random variables and make the first $(k-1)$ of them parents of the kth variable;

2. Randomly select a variable $X_i$ from the rest of the variables and a k-clique from the moral graph of the Bayesian network in the previous step. Make each variable of the selected k-clique a parent of $X_i$;

3. Repeat previous step until all the variables have been considered;

4. For each variable, randomly remove some variables from its parent set.

It is easy to see that the moral graph of the RBNBT has a treewidth at most $k$ with probability one for any problem sizes. Implementation details and basic properties of the RBNET also deserves further exploration.

### Acknowledgements

I would like to thank Professor Joseph Culberson for many helpful discussions and anonymous referees for their comments.